\documentclass[runningheads]{llncs}

\usepackage{csquotes}
\usepackage{graphicx}

\usepackage{amsmath,amsfonts,bm}

\def\eqref#1{equation~\ref{#1}}
\def\1{\bm{1}}

\DeclareMathAlphabet{\mathsfit}{\encodingdefault}{\sfdefault}{m}{sl}
\SetMathAlphabet{\mathsfit}{bold}{\encodingdefault}{\sfdefault}{bx}{n}

\usepackage[utf8]{inputenc} % allow utf-8 input
\usepackage[T1]{fontenc}    % use 8-bit T1 fonts
\usepackage{hyperref}       % hyperlinks
\usepackage{url}            % simple URL typesetting
\usepackage{booktabs}       % professional-quality tables
\usepackage{amsfonts}       % blackboard math symbols
\usepackage{nicefrac}       % compact symbols for 1/2, etc.
\usepackage{microtype}      % microtypography
\usepackage{graphicx}
\usepackage{color}
\graphicspath{ {./images/} }
\usepackage{adjustbox}

\begin{document}
\title{Critic Guided Segmentation of \\ Rewarding Objects in First-Person Views}

\author{Andrew Melnik\inst{1}* \and
Augustin Harter\inst{1}* \and
Christian Limberg\inst{1} \and\\
Krishan Rana\inst{2} \and
Niko Sünderhauf\inst{2} \and
Helge Ritter\inst{1}
}

\authorrunning{A. Melnik et al.}
\institute{CITEC, Bielefeld University, Germany \and
Centre for Robotics, Queensland University of Technology (QUT), Brisbane, Australia\\
\email{andrew.melnik.papers@gmail.com}\\
\email{aharter@techfak.uni-bielefeld.de}
}
\maketitle
\begin{abstract}
This work discusses a learning approach to mask rewarding objects in images using sparse reward signals from an imitation learning dataset. For that, we train an \textit{Hourglass} network using only feedback from a critic model. The \textit{Hourglass} network learns to produce a mask to decrease the critic's score of a high score image and increase the critic's score of a low score image by swapping the masked areas between these two images. We trained the model on an imitation learning dataset from the NeurIPS 2020 MineRL Competition Track, where our model learned to mask rewarding objects in a complex interactive 3D environment with a sparse reward signal. This approach was part of the 1st place winning solution in this competition. Video demonstration and code: \url{https://rebrand.ly/critic-guided-segmentation}
\end{abstract}

\let\thefootnote\relax\footnote{*Shared first authorship}

% TODO @ andrew: RL vision
% TODO @ anyone: fill discussion part
% TODO @ anyone: apply correct template
% TODO @ anyone: more introduction paragraphs?

\section{Introduction}

Training a semantic segmentation network can be a difficult problem in the absence of label information. We propose using sparse reward signals from a Reinforcement Learning (RL) environment to train a semantic segmentation model for masking rewarding objects. Moreover, our approach allows training the semantic segmentation model entirely on an imitation learning dataset without interaction with the environment. This is of interest for a number of use cases. Such technique can contribute to explainable AI \cite{gunning2019darpa}, symbolic and causal reasoning in the space of detected objects, robotics, or as auxiliary information \cite{jaderberg2016reinforcement} for an RL setup. 
Learning the optimal policy in sparse reward environments \cite{guss2021towards} is an important challenge in Deep Reinforcement Learning (DRL) \cite{bach2020learn,10.3389/frobt.2021.538773,harter2020solving}. A masking network for rewarding objects can support an actor-critic RL setup in a way that is intuitive and explainable to a human. Semantic segmentation of rewarding objects can aid transfer learning, better sample efficiency of training, and better performance of a trained agent. In the following, we are considering how a trained critic model - value network trained on discounted reward values - can guide the direct segmentation of such rewarding objects in a feed-forward manner. 

\begin{figure}
  \centering
%   \vspace{-5mm}\hspace*{-6mm}\includegraphics[scale=0.2]{images/fig-results.png}
  \includegraphics[scale=0.109]{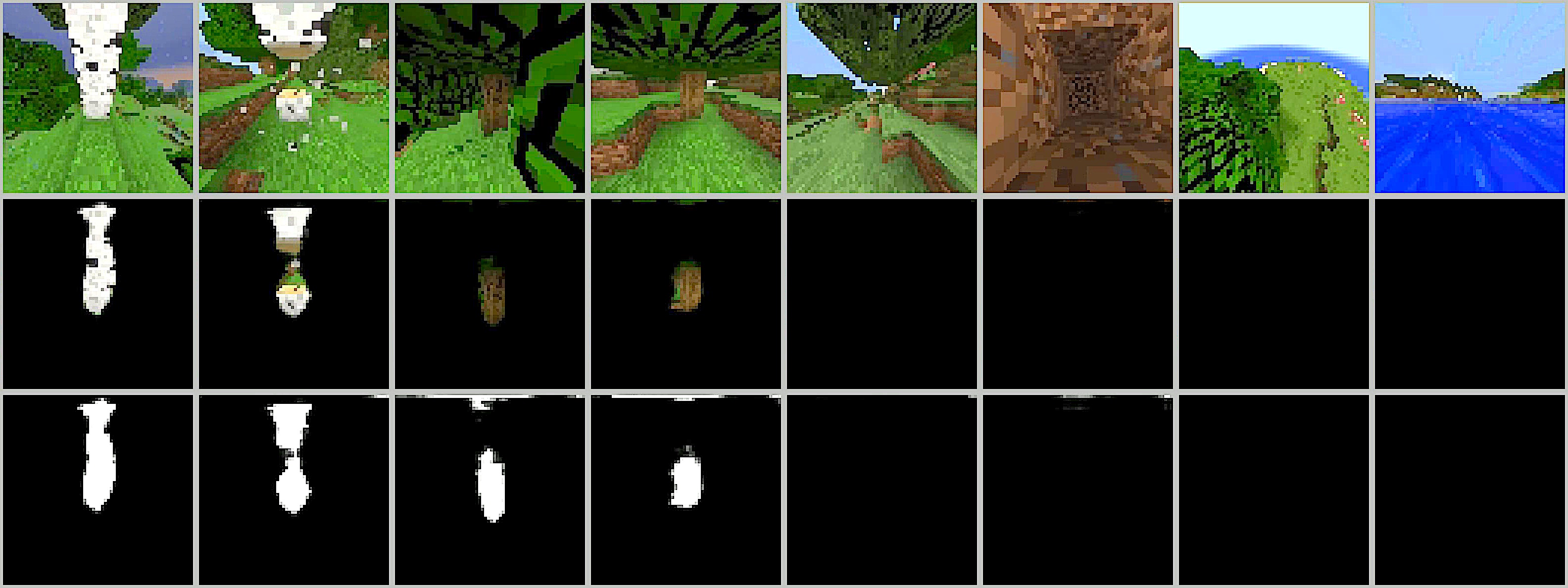}
  \caption{Segmentation results: The \textit{Hourglass} model learns to segment rewarding objects (tree trunks) without any label information but only from reward signals. In the first four columns, showing high critic-value images, the trained \textit{Hourglass} model detects different instances of rewarding objects (white and brown tree trunks). The model is resistant to generation of false-positive masks in low score images (columns 5-8). The first row shows the input images, the second row shows the segments extracted from the input images using the generated masks (not ground truth), and the third row shows the masks generated by the \textit{Hourglass} model. Video demonstration and code: \url{https://rebrand.ly/critic-guided-segmentation}}
  \label{fig-results}
\end{figure}

To this end, we use the critic to select two sets of images characterized by high vs. low expected reward values. Using these two sets of images from an imitation learning dataset, we can train a segmentation network for rewarding object regions by applying it to high expected reward images and exchanging the masked region with the low expected reward image. This makes it possible to evaluate how the mask changes the reward value prediction and therefore allows us to train the segmentation network to mask the rewarding objects (Figure \ref{fig-results}, see Section \ref{sec:experiments} for details).

This work was motivated by the \enquote{MineRL NeurIPS 2020 Competition: Sample Efficient Reinforcement Learning in Minecraft} \cite{guss2021towards}, where our AI agent won the 1st place. See Section \ref{sec:experiments} for details on the challenge, goals, and objects that an agent needs to learn in the world of Minecraft. While our approach is formulated in a domain-agnostic way, we evaluate it on first-person views in 3D Minecraft environments from the NeurIPS MineRL 2020 Competition. 

% \niko{Why do we want that? What is the advantage of doing this? Are you interested in better semantic segmentation? Better performance of your trained agent? Better sample efficiency? Transfer? ... Important to frame the paper clearly and set the expectations.} 

\begin{figure}
  \centering
  \includegraphics[scale=0.255]{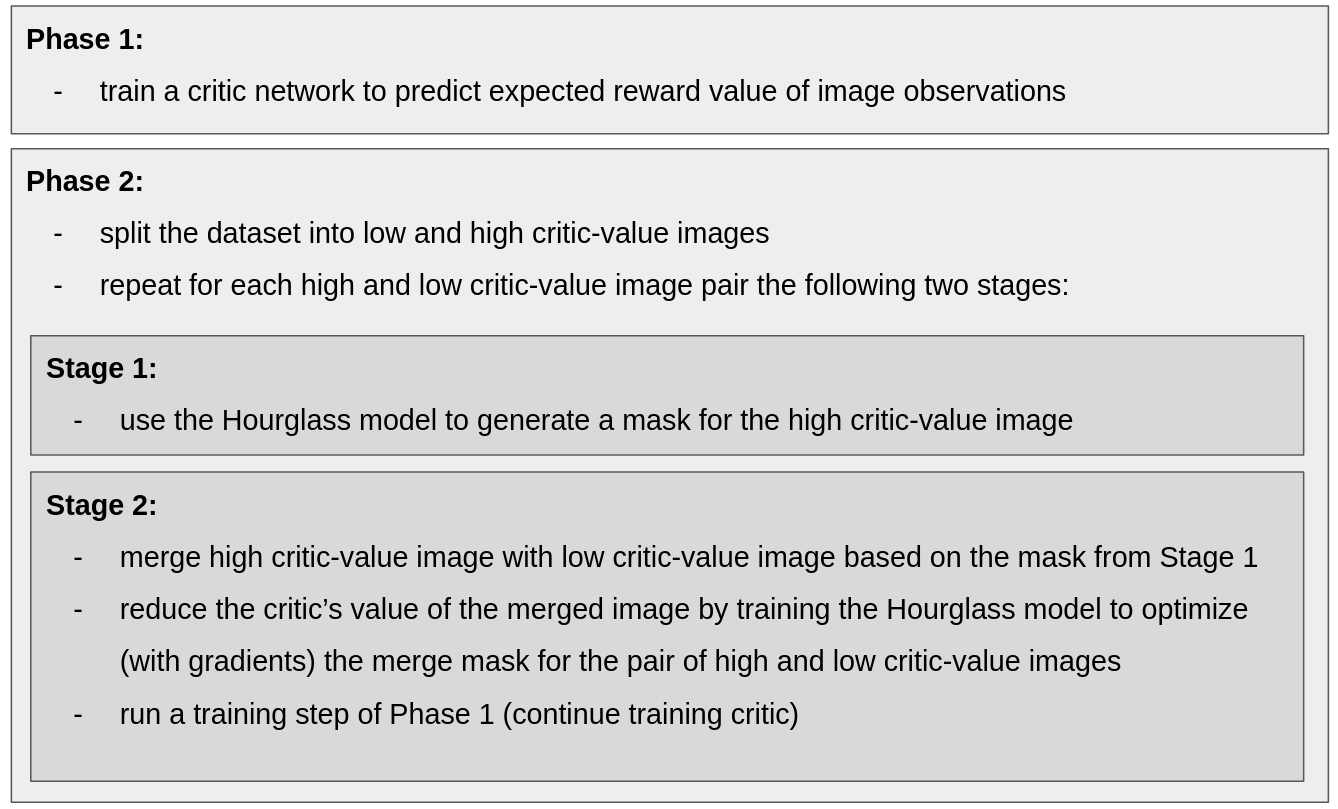}
  \caption{Training pipeline overview. Phase 2 is explained in more detail in Figure \ref{fig-unet}.}
  \label{pseudo-code}
\end{figure}

\begin{figure}
  \centering
  \includegraphics[scale=0.33]{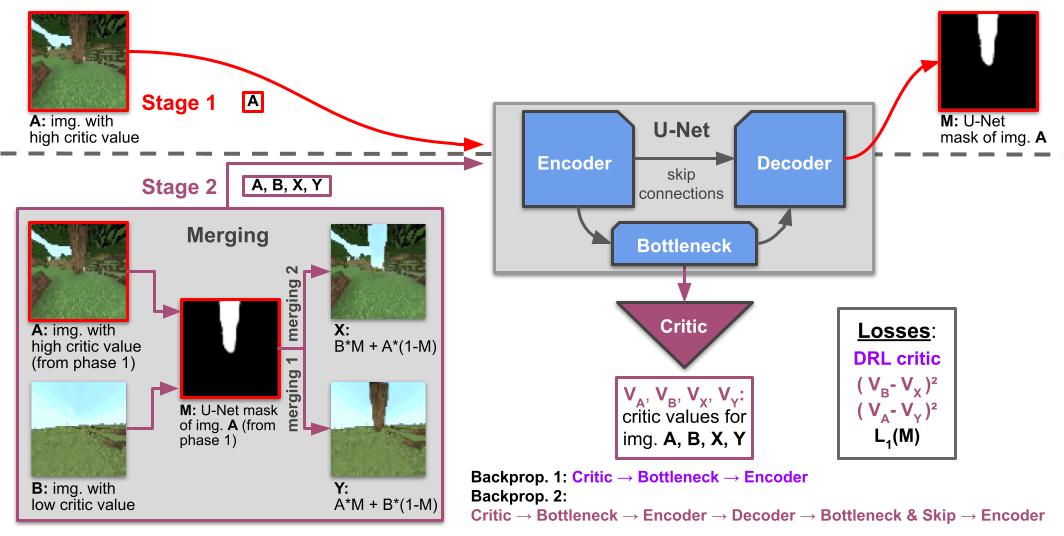}
  \caption{This figure shows the second phase of our pipeline containing the segmentation training, which consists of two stages: Stage 1 (highlighted in red): Image \textbf{A} (high critic value) passes through the \textit{Hourglass network}, forming a mask \textbf{M}. Stage 2: the mask \textbf{M} is used to merge image \textbf{A} (high critic value) with image \textbf{B} (low critic value) resulting in image \textbf{X} (masked parts of \textbf{A} replaced with \textbf{B}) and image \textbf{Y} (masked parts of \textbf{A} injected in \textbf{B}). Images \textbf{A}, \textbf{B}, \textbf{X}, and \textbf{Y} are then passed through the encoder and critic. The losses penalize differences in critic values for image pairs \textbf{A} : \textbf{Y}, and \textbf{B} : \textbf{X}. A linear regularization term penalizes mask intensity and prevents collapse to a trivial solution where the mask \textbf{M} fills the entire image. Video explanation and code: \url{https://rebrand.ly/critic-guided-segmentation}}
  \label{fig-unet}
\end{figure}

\newpage
\section{Methods}

Refer to Figure \ref{pseudo-code} to see a high-level overview of our approach, which we will now describe in more detail: We first train a critic model to predict the discounted reward for images and then train an \textit{Hourglass} model \cite{hourglass} to infer a segmentation mask over rewarding objects while continuing to train the critic (Figure \ref{fig-results}). The architecture is based on two sub-modules (Figure \ref{fig-unet}): A critic network trained to predict discounted reward for a given image, and an \textit{Hourglass} model that produces a segmentation mask for rewarding objects in the image. Such masks, when used to replace parts from high critic score images, lowers the predicted score and, when used to inject these parts into low critic score images, increases the predicted score. To achieve this objective, the \textit{Hourglass} model learns to segment the rewarding objects.

We evaluate our approach in the first-person Minecraft environments from the \textit{MineRL 2020 Competition} \cite{guss2021towards}. The provided imitation learning database consists of data recorded from human players, and we use only images and the recorded sparse reward signals from the database to train our model.

\subsection{Pipeline}
% ndrw: replace phase 1 and phase 2 with critic training and mask training ?
% ndrw: critic picks up the effect of the mask
The training is divided into two phases: \textit{Phase 1} contains the initial critic training enabling us to use the critic in \textit{phase 2} to train our segmentation model. \textit{Phase two} happens in two stages and is shown in Fig \ref{fig-unet}: \textit{Stage 1} is the pass through the segmentation model producing a segmentation mask. We then use this mask in \textit{stage 2} to construct new merged images and pass them through the critic to calculate the gradient with respect to the mask-based merging.
% TODO \cla{a mixing of paragraphs and subsubsections looks weird, I would rather stick to paragraphs. Also after a paragraph title should be a . or a : at the end (maybe write it like "Phase 1 (Initial Critic Training):" to be consistent)}
\subsubsection{Phase 1: Initial Critic Training} We train the critic model to directly predict the time discounted reward value of states which are 64x64x3 RGB single image observations. We use the time discounted reward from the data set episodes as a supervised training signal together with the mean squared error as a loss function. After training converges, we use the critic to split the database into images with high critic values \textbf{A} and low critic values \textbf{B} for \textit{phase 2}.

\subsubsection{Phase 2: Segmentation Training}\textit{Phase 2} is subdivided into two stages and visualized in Fig \ref{fig-unet}): In \textit{stage 1} we pass an image through the segmentation model to produce a mask. In \textit{stage 2} we use this mask to swap out pixels in an image pair and pass the resulting merged images through the critic to infer how the mask-based pixel swap changed the original critic value prediction for the image pair. 
We do this because we do not have explicit ground truth masks segmenting the rewarding objects as a training target. Instead, we pick image pairs such that one (image \textbf{A}) has a high critic value and the other (image \textbf{B}) a low critic value and use these pairs to formulate a loss which enables learning to segment rewarding objects: The key idea is that reward related parts of images should decrease the critic value when removed and increase the critic value when injected into another image.

\paragraph{\textbf{Implementation:}} We implement this idea by first using the segmentation model in \textit{stage 1} to produce a mask \textbf{M} based on the high critic value image \textbf{A}. Next, in \textit{stage 2} we use the mask to swap the highlighted pixels in \textbf{A} and replace them with the corresponding pixels in low critic value image \textbf{B}. This generates a second pair of images: \textbf{X} (image \textbf{A} with masked pixels replaced by content of \textbf{B}) and \textbf{Y} (image \textbf{B} with content substituted by masked pixels from image \textbf{A}). For a \enquote{perfect} mask that captures all reward related structures, the critic values between the pairs should be swapped: High in \textbf{Y} since it received all reward related content of \textbf{A}, and low in \textbf{X} since all reward-related content has been replaced by contents from low-value image \textbf{B}). The \textit{inject loss} penalizes the squared difference between critic values of \textbf{A} and \textbf{Y}: $L_R = (V_A-V_Y)^2$ And the \textit{replace loss} penalizes the squared difference between critic values of \textbf{B} and \textbf{X}: $L_I = (V_B-V_X)^2$

To further illustrate this, we can consider a bad mask that leaves all essential reward elements in \textbf{X} contributing to a high \textit{replace loss}, since then $V_X$ remains high, while $V_B$ is low; similarly, \textbf{Y} would hardly receive any reward related content from \textbf{A}, keeping $V_Y$ low whereas $V_A$ is high and therefore producing a high \textit{inject loss}.

\paragraph{\textbf{Regularization:}} In order to avoid trivial solutions like a full image mask replacing the complete images, we apply a linear regularization to enforce minimal masks: $L_N = |M|$

\paragraph{\textbf{Continued Critic Training:}} Training the mask has an influence on the encoder weights, which are shared across critic and segmentation models and therefore can mess up the critic predictions. A straightforward solution would be to freeze the encoder weights after the initial critic training in \textit{phase 1} is over. This works, but we can allow the segmentation model to influence the encoder weights if we also continue to train the critic like in \textit{phase 1}, which keeps the encoder functional for reward prediction. This increases performance; see \textit{frozen weights} for comparison in Section \ref{sec:experiments}.

\paragraph{\textbf{Handling False Positives:}}
So far, we have explained how we pass high critic value images through the segmentation model to produce reward related segmentation masks. However, this means that the model would only see high-value images and therefore be heavily biased towards producing a mask even for low reward images producing \enquote{false positives}. Therefore we take low-value images for \textbf{A} in half of the time. This means that a batch containing 128 images would contain 32 high-value and 32 low-value images as \textbf{A} and 64 low-value images for \textbf{B}.
With this, we can keep the above-stated losses without producing \enquote{bad} gradients since $(V_A-V_Y)^2$ and $(V_B-V_X)^2$ are small because both images have similar (low) critic values already before merging. This allows the gradient from the regularization term that favors sparse masks to drive the segmentation response towards the desired empty mask output when receiving a low critic image as input. Segmentation results for both high and low-value images are shown in \ref{fig-results}.

\section{Experiments}
\label{sec:experiments}

\paragraph{\textbf{Data Set:}} We apply our method on an imitation learning data set from the NeurIPS2020 MineRL Challenge. It is based on Minecraft, a game providing a 3D world where players can interact with the world in the first-person view. We used the  \textit{TreeChop} environment, which contains episodes of human players chopping trees and collecting wood: The players can repeatedly use the \textit{attack action} on a tree trunk to destroy it which produces the item \textit{log} which is automatically collected when in proximity. Only upon collection the player receives \textit{reward = 1}; all other images give \textit{reward = 0}. Less than 1\% of images get a reward signal, while roughly a third of the images contain views of trees in close proximity and almost all images contain some tree features in sight but possibly further away.

After chopping the base trunk of a tree, players usually stand below the \enquote{floating} tree crown to chop the remaining wood in the tree crown. We remove most of this tree crown chopping which makes up roughly 20\% of images, to focus the reward signal on approaching and chopping the tree trunks. This is automatically done by removing the 35 images after a reward signal. From that, we assemble a data set containing 100k images where we clip reward values higher than 1 (when collecting multiple \textit{log items} at once). We then use a factor of 0.98 to discount the reward every time step and use the resulting discounted reward as the training label for that image. This results in a data set with a quite balanced histogram of discounted reward values, meaning that there are roughly equal amounts of images for every reward value in the range from 0 to 1, which stabilized the critic training. Additionally, we apply a small data augmentation that shifts the images randomly up to 12 pixels to the right or left, which works against the strong bias in the data set that trees are almost always in the middle of the image when receiving a reward signal.

% TODO \cla{cite hourglass paper}
\paragraph{\textbf{Architecture}}We use a convolutional encoder-decoder architecture with skip connections inspired by \textit{Hourglass Networks}, which has two outputs: The first output is the critic score estimating the value of images. Its implemented through two additional linear layers after the decoder bottleneck (Figure \ref{fig-unet}). The second output is the segmentation mask which is produced by the decoder. Our simple custom-made network has an encoder with 5 convolution layers with 40, 40, 40, 80 and 160 channels respectively, and kernel size 3x3. The last layer results in a non-spatial bottleneck (dimensions: 1x1x160). Each layer is followed by a LeakyReLU and we use max pooling after the first 4 layers. The decoder has a mirrored structure, but we switch the pooling layers with upsampling. Its output layer is passed through the Sigmoid function to produce the mask. The critic shares the encoder, and after the bottleneck additionally consists of two fully connected layers with 160 units each. Further, we use a 50\% dropout after the third and fourth encoder layers and after the first fully connected critic layer.

Reward prediction and segmentation demand similar features, so instead of using separate models they both share the encoder: In the first phase, the encoder is trained to create a meaningful feature representation of images that makes it possible for the critic to predict the reward. In the second phase, the decoder can use skip connections to access the encoder representation, which we found greatly improves mask quality in comparison to a separate critic. See Section \ref{sec:results}.

\paragraph{\textbf{Evaluation:}} To test the performance of our model, we collect a test set containing 18k images from 10 episodes of a player chopping tree trunks. We modified the environment simulator enabling us to extract the ground truth segmentation masks of tree trunks; such masks were never seen during training. With this, we can measure the \textit{Intersection over Union} (IoU) score as a performance metric to evaluate our segmentation model. We report the performance in Table \ref{tab:iou} and compare it to a saliency map baseline.

% TODO \cla{stay consistent with capitalization of e.g. Section, Table, Figure. Wenn man ueber eine bestimmte Section/Figure/Table spricht (in Table 1 ...), dann kann man es groß schreiben (also eigentlich fast immer), ansonsten klein. Oder einfach immer klein geht auch. (cref package is quite good for this (maybe consider for the next paper ;) ))}

\paragraph{\textbf{Training Details:}} We train the critic in \textit{phase 1} on the 100k images for 15 epochs using a batch size of 64 until convergence and use it to split the data set into images with a predicted reward higher than 0.7 (resulting in 20k images) and images with predicted reward lower than 0.3 (resulting in roughly 30k images). These values were obtained through a hyper-parameter grid-search with the evaluation dataset. We then use these two subsets to continue our training in \textit{phase 2} for one epoch with a batch size of 128.

\section{Results}
\label{sec:results}

We report the intersection-over-union (IoU) scores of the segmentation mask achieved when compared to the ground truth data. To better understand how our model works, we compare four different model variants with two baseline approaches which we describe in detail below before presenting our final results in Table \ref{tab:iou}. Further, we report the performance when post-processing with Conditional Random Fields (CRF) \cite{CRF,CRFrepo}, which is a common method to improve inaccurate or noisy segmentation masks. We additionally provide some visual segmentation results attained by our approach in Figure \ref{fig-results} for high and low critic score images. Once again, these masks are learned only from a sparse reward signal in the challenging 3D Minecraft environment with different lighting conditions, tree types, and tree colors.

\begin{table}
    \centering
    \begin{tabular}{lc}
    \toprule
    \textbf{Model Variant} & \textbf{IoU} \\
    \midrule
    Baseline Full Mask & 0.12 \\
    Baseline Saliency Map & 0.22 \\
    Baseline Saliency Map + CRF & 0.11 \\
    \midrule
    Separate Critic & 0.27 \\
    No Inject Loss & 0.35\\
    Frozen Encoder & 0.38 \\
    Full Model & 0.41 \\
    \textbf{Full Model + CRF} & \textbf{0.45} \\
    \bottomrule
    \end{tabular}
    \vspace*{3mm}
    \caption{\textit{IoU} mean value over 10 training seeds.}
    \label{tab:iou}
\end{table}

\paragraph{\textbf{Baseline Full Mask:}} Full mask covering everything in every image. We report this value for better comparison and interpretation of the other scores. This value can also be interpreted as the percentage of ground truth tree trunk pixels in our test set.

\paragraph{\textbf{Baseline Saliency Map:}} Following \cite{saliency} we compute the Jacobian of the input image with respect to the critic's prediction and weight the Jacobian of each image with the critic's predicted score. We then produce a mask by thresholding each pixels weighted gradient based on a value that is a multiple of the mean pixel gradient values. The exact multiple is determined through a hyper-parameter search. This method can act as a baseline for the task, since calculating saliency maps to visualize where a model is \enquote{looking} at is common practice and requires no training of an additional model. However, the resulting masks are noisy and do not allow to focus on the area of the rewarding object in the image.

\paragraph{\textbf{Baseline Saliency Map + CRF:}} Post-processing the saliency maps with CRF. The CRF hyperparameters were obtained through a grid search with the test set. For some images, it leads to a improved segmentation but overall the saliency maps are too chaotic and noisy for this method to work properly.

\paragraph{\textbf{Separate Critic:}} To test whether our hypothesis is valid that the encoder features learned during critic training are also useful for the segmentation training and therefore motivated the idea of sharing the encoder, we trained a model with a separate encoder for the critic. This means that a separate encoder has to be learned from scratch during segmentation training. The resulting decrease in IoU performance strengthens our hypothesis and favours weight sharing.
 
\paragraph{\textbf{No Inject Loss:}} This model is trained without the \textit{inject loss}, only using the mask to replace parts of high reward images and not injecting them into low reward images. The performance results show a clear decrease in performance, emphasizing the usefulness of the inject objective. It seems like only having to decrease the critic value can lead to more degenerate solutions instead of also having to increase the critic value with the same mask.

\paragraph{\textbf{Frozen Encoder:}} Here we train the model with frozen encoder weights, since the segmentation gradients will influence the encoder and through that mess up the critic predictions needed for our replace and inject losses. With a frozen encoder, the decoder can access the encoder features but not change them to better fit the segmentation task.

\paragraph{\textbf{Full Model:}} The alternative to a frozen encoder is to let the decoder gradients influence the encoder but at the same time continuing to train the critic which prevents the encoder features to become dysfunctional for the critic's reward prediction. This setup results in the best-performing variant of our approach without post-processing. It combines the inject loss with a shared encoder and continued critic training in phase 2 into our \enquote{Full Model}.

\paragraph{\textbf{Full Model + CRF:}} Post processing the model output with CRF. The CRF hyperparameters were obtained through a grid search with the test set. In contrast to the saliency maps, here, the CRF improves performance.

\section{Discussion}

Humans explicitly learn notions of objects. There has been extensive research inspired by psychology and cognitive science \cite{melnik2018world,konig2018embodied,konen2019biologically} on explicitly learning object-centric and reward-centric representations from pixels \cite{simonyan2013deep}.
Focusing on the reward quality of objects also adds an interesting perspective on the question of how NN build up their understanding of images. These questions have been studied by feature visualization methods \cite{olah2017feature} and interpretability techniques \cite{olah2018building}.

Understanding RL Vision is a challenging problem. Hilton et al. \cite{hilton2020understanding} proposed to analyze, diagnose and edit deep reinforcement learning models using attribution. Although this technique allows highlighting objects related to positive and negative rewards in 2D environments, it requires human domain knowledge to select the proper attributions, and it was not shown that this technique will work in more complicated 3D first-person-view environments. To understand how an agent learns and executes a policy, a method for generating salience maps was introduced \cite{greydanus2018visualizing}. In this method, a change in the value function was used when sampling a grid of perturbations.

It has been empirically observed that RL from raw pixels is sample-inefficient \cite{kaiser2019model,melnik2019modularization}. Learning policies from state-based features is significantly more sample-efficient than learning from pixels. An approach called CURL \cite{srinivas2020curl} shows that by extracting high-level features from raw pixels using contrastive learning and then using these features as state in the RL setting results in a superior sample efficiency during training of an RL agent. In contrast, our approach allows learning of explicit masks over rewarding objects. Extracting these segmentation masks which are derived directly from reward can help prepossess the most relevant information from the raw pixel state representation. This is of interest for a number of use cases in explainable AI, symbolic and causal reasoning \cite{melnik2019combining} in the space of detected objects, robotics, as well as in RL setups. Moreover, our approach can learn the segmentation model solely from imitation learning dataset. 

DRL has been applied successfully in various domains for training high-performing agents \cite{schilling2018approach}. Our approach could be used as an auxiliary module to train the agent from demonstrations to improve sample efficiency of learning. Our contribution is a novel and an intuitive use of joint training of a critic network and an image segmentation approach to highlight rewarding object segments in reinforcement learning environments. In this contribution, we showed that it is possible to train a model to generate high-quality masks depicting rewarding objects in images without explicit label information, but only using feedback from the critic model. Our approach was part of the 1st place winning solution in the \enquote{MineRL NeurIPS 2020 Competition: Sample Efficient Reinforcement Learning in Minecraft}. 

Future work may include further experiments with RL or imitation learning setups extended with our model that provides rewarding-object masks. Identifying a mask or heatmap for negative rewards as well as non-reward entities may be a possible continuation of this work. Self-supervised learning of embedded classification of reward-centric objects can facilitate development of causal and symbolic reasoning models.

\bibliographystyle{splncs04}

\bibliography{iclr2021_conference}

\end{document}